\documentclass[10pt, twocolumn]{article}
\usepackage[a4paper, total={6.6in, 10in}]{geometry}%

\title{\LARGE \bf
Optimisation of Body-ground Contact for Augmenting Whole-Body Loco-manipulation of Quadruped Robots
}

\author{Wouter J. Wolfslag, Christopher McGreavy, Guiyang Xin, Carlo Tiseo, \\Sethu Vijayakumar and Zhibin Li\thanks{This work was supported by EPSRC UK RAI Hub for Offshore Robotics for Certification of Assets (ORCA, grant reference EP/R026173/1), the EPSRC CDT in Robotics and Autonomous Systems (EP/L016834/1) and EPSRC UK RAI Hub NCNR (EPR02572X/1) }\thanks{Authors are with the School of Informatics, University of Edinburgh, UK.}
}

\usepackage{cite}
\usepackage{placeins}
\usepackage[inline]{enumitem}
\usepackage{amsmath,amssymb,amsfonts}
\usepackage{algorithmic}
\usepackage{graphicx}
\usepackage{textcomp}
\usepackage{xcolor}
\usepackage{booktabs}
\usepackage{siunitx}
\def\BibTeX{{\rm B\kern-.05em{\sc i\kern-.025em b}\kern-.08em
    T\kern-.1667em\lower.7ex\hbox{E}\kern-.125emX}}
\usepackage[normalem]{ulem}
\usepackage{todonotes}
\usepackage{subcaption}
\usepackage[abs]{overpic}
\usepackage{fontawesome}
\usepackage{balance}

\newcommand{\ti}[1]{_\text{#1}}
    
\newcommand{\prong}{prong}
\newcommand{\prongs}{prongs}

\date{}

\begin{document}

\maketitle
\thispagestyle{empty}
\pagestyle{plain}

\begin{abstract}
Legged robots have great potential to perform loco-manipulation tasks, yet it is challenging to keep the robot balanced while it interacts with the environment.
In this paper we study the use of additional contact points for maximising the robustness of loco-manipulation motions.
Specifically, body-ground contact is studied for enhancing robustness and manipulation capabilities of quadrupedal robots.
We propose to equip the robot with prongs: small legs rigidly attached to the body which ensure
body-ground contact occurs in controllable point-contacts.
The effect of these prongs on robustness is quantified by computing the Smallest Unrejectable Force (SUF), a measure of robustness related to Feasible Wrench Polytopes.
We apply the SUF to assess the robustness of the system, and propose an effective approximation of the SUF that can be computed at near-real-time speed.
We design a hierarchical quadratic programming based whole-body controller that controls stable interaction when the prongs are in contact with the ground.
This novel concept of using prongs and the resulting control framework are all implemented on hardware to validate the effectiveness of the increased robustness and newly enabled loco-manipulation tasks, such as obstacle clearance and manipulation of a large object.
\end{abstract}

\section{Introduction}
\label{sec:introduction}

Combined locomotion and manipulation tasks are a key competence for legged robots in applications such as warehousing, search and rescue, and offshore inspection and maintenance.
To manipulate objects, a robot must exert forces onto the environment. To locomote, the robot must remain balanced and stable under the load of the manipulation.
The main challenge of loco-manipulation is performing these tasks simultaneously by managing the limited resources required to complete them: motor torques and tangential contact forces ~\cite{farnioli2015optimal, abi2019torque}.
Better management of these resources will improve the robot's workspace, payload, robustness and stability.
This paper investigates how to improve that management by adding contact points to a quadruped robot. 

Previous work has shown that extra contact points reduce resource consumption and improve stability. Examples are found in humans or humanoid robots using their arms for balance and in multi-finger and arm manipulation, \cite{kudruss2015optimal}, and \cite{henze2016passivity,carpentier2018multicontact,koolen2016design}, and \cite{smith2012dual} respectively.
Accurate force control at multiple contacts increases robustness of quadrupedal locomotion, especially in rough terrains and with disturbances~\cite{xin2018model}. 

Additional contacts, however, also produce challenges in control, due to the uncertainty in estimating exact contact locations, and dealing with non-trivial surface geometries of the contacting body.
Complex contacts do not fit well into multi-contact frameworks, which rely on simple contact geometry, and often on contacts only occurring at the end of the kinematic chain.
This can limit the versatility of using extra contact points such as knee-ground contact \cite{henze2017multi}, sliding \cite{trkov2019bipedal} or rolling interactions in humanoid robots~\cite{specian2018robotic}.
Recent machine learning approaches address more complex contact scenarios, such as in hand-manipulation and Jenga\cite{van2015learning,fazeli2019see} but also have limited versatility due to challenges of learning.

\begin{figure}[t]
    \centering
    \includegraphics[width=\columnwidth]{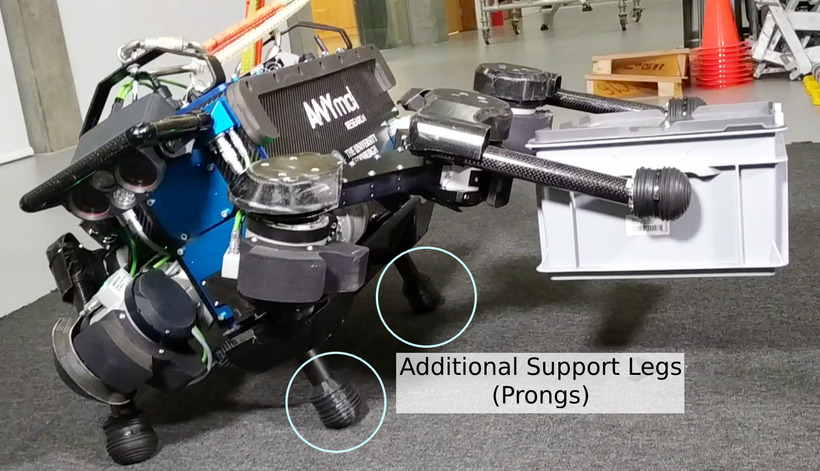}
    \caption{Body-ground contact for enabling diverse loco-manipulation tasks: manoeuvring objects by legs.}
    \label{fig:setups}
\end{figure}

In contrast, humans and animals use various parts of their bodies to increase movement stability.
We are motivated to investigate how quadrupedal robot movement might benefit from additional non-conventional body-ground contacts, which will be evaluated in this paper. 

To enable versatile body-ground contact, we equip a quadruped robot with additional fixed limbs (see Fig. \ref{fig:setups}), which we call \prongs.
These prongs are rigidly attached to the base of the robot and ensure point-contact at a known location.
Such contact fits into the whole-body force control pipeline shown to be versatile in other multi-contact scenarios~\cite{hutter2014quadrupedal}.
Rigidly connecting the prongs to the robot's torso means they will reduce actuator loads by supporting the robot's weight when they are in contact with the ground, thereby allowing the robot to perform additional tasks.
Note the contrast between prong-ground and belly-ground contact: by using prongs, we know the exact contact locations, which would be difficult to estimate when using the belly. Furthermore, the height of the prongs allows the body to be mobile while maintaining contact with the ground, which would be more difficult with belly-ground contact.

Prong-like concepts are seen in wheeled platforms.
Using outriggers, a wheeled robot can resist more disturbance force \cite{shirafuji2016mechanism} \cite{kiyota20063d}, with an estimate of the benefits found in \cite{jeng2010outrigger}.
Legged robots can be augmented with wheels or skates at the feet to speed up locomotion in easy terrain~ \cite{oh2017technical, bjelonic2018skating}, or with a tail to counteract inertial shifts during fast locomotion \cite{saab2018robotic}. 
Augmentations proposed in this paper can be used in parallel with those mentioned above.

While the \prongs~provide controllable ground contact, three open questions remain: how to design \prongs~ so they provide maximal benefit, how to deal with the control challenges posed by body-ground even with the simplified point contacts, and how to plan motions while deciding if and how to make contact with the \prongs.
The focus of this paper is proof-of-concept and analysis of enabling body-ground contact, therefor it considers neither planning, nor further mechanical enhancement such as retractable prongs.

We first deal with prong design, which must consider placement, ground-clearance (length), manipulability and disturbance rejection capabilities.
We define the Smallest Unrejectable Force (SUF) as a metric to quantify disturbance rejection ability in loco-manipulation tasks under actuation limits and interaction constraints, and provide a fast-to-compute approximation, which are used optimise the design of the \prongs.
Our approach extends previous work showing the importance of optimizing posture for robustness \cite{thibodeau2006static,abi2019torque}.

To control the robot with prongs, we use the established framework for modern quadruped robots: Quadratic Programming (QP) based inverse dynamic controllers~\cite{hutter2014quadrupedal, del2016robustness,xin2018model,escande2014hierarchical}.
However, these controllers were designed for contacts at the ends of the kinematic chain, not at the torso.
The control of non-end-effector limb contact has been studied in manipulation \cite{bicchi1993force} considering contacts with moving obstacles which do not kinematically constrain the contact limbs.
When using prongs, the torso will be constrained, and the above controllers might become unstable.
We propose a hierarchical QP controller that uses body-ground contact constraints to minimise motor torques. 

\subsection{Contributions}

Our paper studies the design of prongs for body-ground contact in quadrupedal robots.
We validate their performance in three hardware experiments: push-rejection, obstacle clearance and object manipulation. 
The last two experiments use two conventional legs freed for manipulations by the support of the prongs.
This provides the following contributions:
\begin{enumerate}
    \item A proof-of-concept \prong~design for the ANYmal robot which enables effective body-ground contact (Sec. \ref{sec:prong_design}).
    \item A novel method to quickly compute an approximation of the Smallest Unrejectable Force, a measure for the robustness of the robot (Sec. \ref{sec:polytopes}).
    \item Metrics for benchmarking the robustness and stability of a robot with and without prongs (Sec. \ref{sec:experiments}).
    \item A hierarchical QP controller that enables the robot to be operated with prongs by including contact constraints on base movement (Sec. \ref{sec:control}).
\end{enumerate}

Section \ref{sec:prelim} explains our notation for robot dynamics.
Section \ref{sec:prong_design} discusses the optimal design of the prongs.
Our robustness measure, the SUF, and novel approximations of it are explained in Section \ref{sec:polytopes}.
Section \ref{sec:simulation_results} shows results from simulations and optimisations.
Section \ref{sec:control} explains our controller for the hardware experiments.
Hardware experiments highlighting the efficacy of the prongs are shown in Sect. \ref{sec:experiments}.
Finally, discussion and conclusion are in Secs. \ref{sec:discussion} and \ref{sec:conclusion}.

\section{Preliminaries/Robot Dynamics}
\label{sec:prelim}

The dynamics of a quadrupedal robot with a manipulator, and \prongs~attached, as shown in Figure \ref{fig:optimal_no_prong}, are given by:
\begin{equation}
    M(q)\ddot{q} + h(q,\dot{q}) {\small=} d(\ddot{q},\dot{q},q) {\small=}  B\tau+J\ti{f}^{\top}\lambda\ti{f}+ J\ti{p}^{\top}\lambda\ti{p}+J\ti{e}^{\top}F, 
    \label{eq:eom}
\end{equation}
where $q$ are the generalised coordinates of the robot describing the position and orientation of the body, and the position of each joint, $M(q)$ is a positive definite mass matrix, $h(q,\dot{q})$ is the dynamic bias containing of centrifugal, Coriolis and gravitational effects, $\tau$ are the joint torques, $B$ is a selection matrix, $J\ti{f}$, $J\ti{p}$ and $J\ti{e}$ are the Jacobians of the feet, prongs and end-effector of the arm respectively, and $\lambda\ti{f}$, $ \lambda\ti{p}$ and $F$ are external (reaction) forces at those points.

These equation of motions are subject to further constraints to ensure physically feasible ground interaction and joint/motor torques.
Ground interaction constraints ensure the robot does not slip or penetrate the ground, and are only considered when the associated body part is in ground contact.
For computational efficiency, these conditions are approximated as linear constraints for each contact point $i$:
\begin{align}
    \begin{bmatrix}0 & 0 -1\end{bmatrix}\lambda_i &\leq 0 \\
    \begin{bmatrix} 1 & 0 & \small{-\frac{1}{2}}\sqrt{2}\mu\end{bmatrix}\text{abs}(\lambda_i) &\leq 0 \\
    \begin{bmatrix} 0 & 1 & \small{-\frac{1}{2}}\sqrt{2}\mu\end{bmatrix}\text{abs}(\lambda_i) &\leq 0\\
    J_i\ddot{q} + \dot{J_i}\dot{q} &= 0
    \label{eq:constraints}
\end{align}
where $\mu$ is a friction coefficient, and the abs-operator returns the piece-wise absolute value.
Additionally, the motor capabilities are reflected in bounds on the joint torque for each joint index $i$ and torque limit $\bar{\tau}_i$:
\begin{equation}
-\bar{\tau}_i \leq \tau_i \leq \bar{\tau}_i. 
\label{eq:torque_bounds}
\end{equation}

\section{Optimal Prong Design}
\label{sec:prong_design}

The prongs enlarge the buffer between the motor torque limits  and the torques required to stand, which can then be used to reject disturbances or perform secondary tasks.  However, the magnitude of the benefits depend on how the prongs are placed and sized. 
To investigate the effects of the \prong~design  we focus on a scenario in which a force is applied at the end-effector of an arm attached to the torso of a robot (Fig. \ref{fig:setups}). We find the effect of \prong~placement on the size of the disturbance the end-effector can sustain without moving the robot.
We then optimise the prong placement and the robot configuration for this robustness measure.

By using a two \prong~configuration, the robot can either fix torso pitch (by grounding both \prongs) or allow pitching (by grounding one \prong) depending on task requirements.
To ensure symmetry both prongs have equal length, and are placed on the $x$-axis of the robot frame. Furthermore, we enforce a symmetric position of the feet.
In this configuration, maximum robustness is achieved when the prongs are furthest apart, so \prongs~are placed as far apart as possible without interfering with the leg motion.
As a result, the optimisation only requires three parameters:
\begin{equation}
\max_{\{x_f, y_f, b_z\}} F_{\text{SUF}}(x_f, y_f, b_z), \label{eq:optimization}
\end{equation}
where the SUF is a measure for the robustness (defined in the next section) and $x_f$, $y_f$ and $b_z$ are the feet $x$ and $y$ position and torso $z$ position.
These decision variables are shown in Figure \ref{fig:optimal_no_prong}.
When optimizing \prong~position, the \prong~length matches the height of the torso.
To solve the inverse kinematics, we use a standard iterative procedure with the transpose Jacobian, which requires no further regularization.

\section{Smallest Unrejectable Force}
\label{sec:polytopes}

A key element of he robustness of a robot is the amount of external force it can withstand while tracking a target motion.
Computing such forces and associated robustness metrics can be done via Feasible Wrench Polytopes, as discussed for legged robots in \cite{orsolino2018application}, and for manipulation in\cite{ferrolho2019comparing}.
The FWP is the set of wrenches applied to the robot, such that the ground reaction forces and joint torques required to execute the desired motion stay within the friction cone and motor limits respectively.
Here, we are interested in a slight variation: the Rejectable Force Polytope (RFP), the set of forces that can be applied to the robot at the end-effector, such that it is able to perform the desired accelerations while satisfying the constraints in Eqs. \ref{eq:eom}-\ref{eq:torque_bounds}.
\begin{align}
 \mathcal{F}_{\text{RFP}}(q,\dot{q},\ddot{q}\ti{d}) = \{F\in\mathcal{R}^3 |& \text{Eqs. \ref{eq:eom}-\ref{eq:torque_bounds} hold for some }\nonumber\\
 &\text{values of} (\lambda_f, \;\lambda_p \;\text{and}\; \tau),\nonumber\\ &\;\ddot{q} = \ddot{q}_d\},
\end{align}
where $\ddot{q}\ti{d}$ is the desired acceleration.
The RFP is a polytope, as the constraints are linear in the free variables.

In practice, it is desirable to summarise the RFP into a single robustness metric.
For this we propose the Smallest Unrejectable Force: the smallest disturbance force the robot cannot withstand while performing its desired motion. 
This is the same as finding the the Chebyshev radius of the RFP, but with the centre of the circle fixed to the origin.

\newcommand{\trimwidthpolytopedown}{0cm}
\newcommand{\trimwidthpolytopeup}{0cm}

\begin{figure}[t]
\vspace{0.2cm}
    \centering
    \begin{subfigure}{0.49\textwidth}
        \includegraphics[trim={0 \trimwidthpolytopedown{} 0 \trimwidthpolytopeup{}},clip,width=\textwidth]{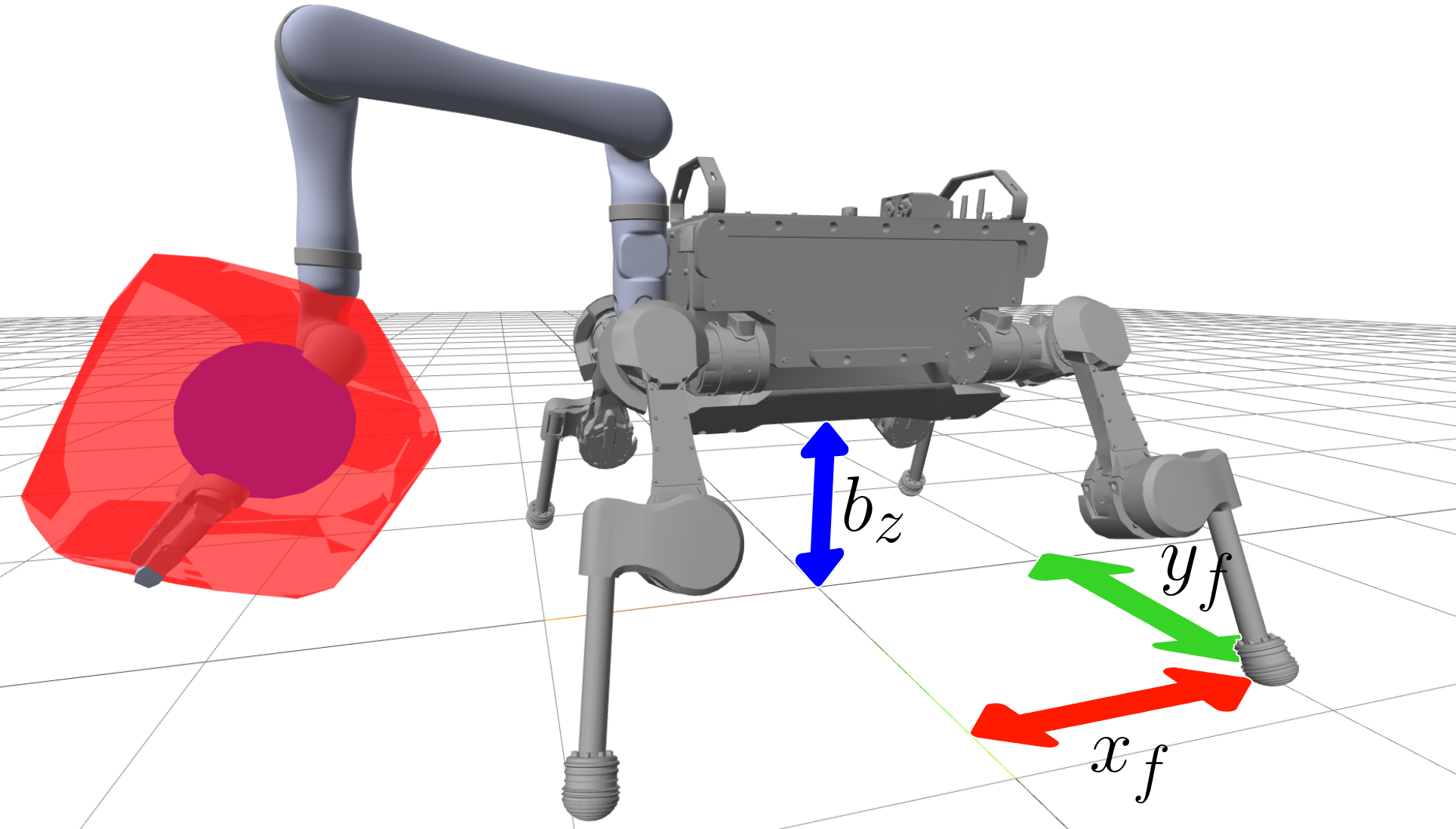}
    \end{subfigure}
    
    \begin{subfigure}{0.49\textwidth}
        \centering
        \includegraphics[trim={0 \trimwidthpolytopedown{} 0 \trimwidthpolytopeup{}},clip,width=\textwidth]{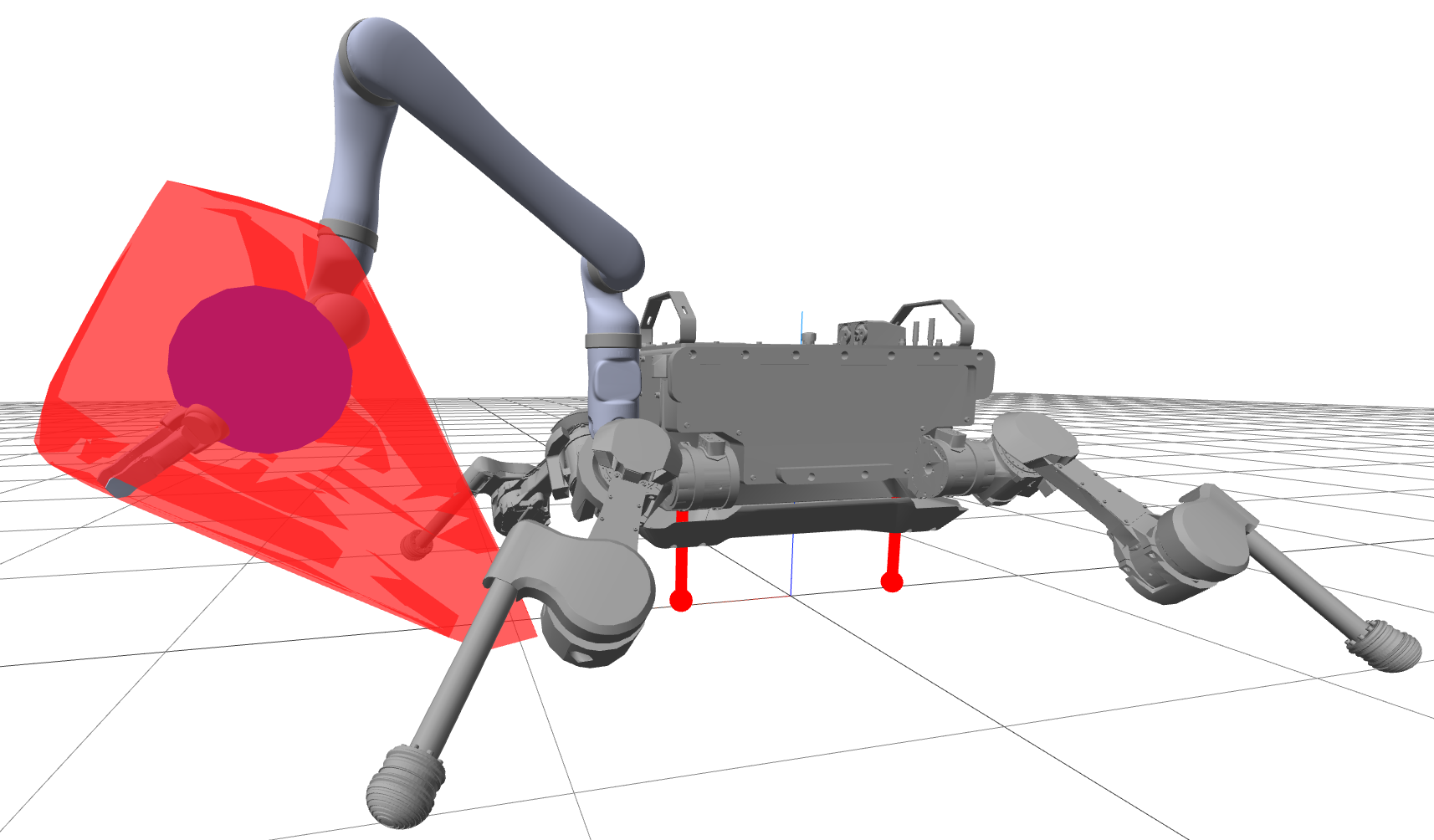}
    \end{subfigure}
    \caption{Rejectable Force Polytope and maximal rejectable force for optimised robot configuration with prong (bottom) and without prongs(top). The end-effector is set to a world frame position: \{0.8m, 0.2m, 0.4m\}.}
    \label{fig:optimal_no_prong}   
\end{figure}

The scheme from \cite{xin2019bounded} can compute the exact RFP,
but its computation time does not scale well with the number of contacts and joints, as it requires a transition between vertex-representation of the FWP to its half-space-representation.
Computing the SUF from the RFP is also computationally expensive, so a simplified metric that finds the Smallest Unrejectable Force in a \text{single} predetermined direction was proposed in \cite{orsolino2018application}.
Here we propose and investigate three approximations of the SUF: Fibonacci, affine, and quadratic.

The Fibonacci approximation is based on an inner (conservative) approximation for the RFP.
Points on the boundary of the RFP are found by solving the optimisation problem:
\begin{equation*}
    \max_{f, \lambda_f, \lambda_p, \tau} \boldsymbol{f} \quad \text{s.t.} \quad  \text{Eqs. \ref{eq:eom}-\ref{eq:torque_bounds}, }
     F = f\hat{F}
\end{equation*}
where the resulting $f$ is the maximum feasible scaling factor for force in the direction $\hat{F}$, considering the dynamic equations and leg joint torque limits $\bar{\tau}$.

The approximation of the RFP is the polytope spanned by vertices found by solving the above optimisation for a set of approximately uniformly distributed force directions according to the Fibonacci-sphere.

To determine the size, $\rho$ of the Smallest Unrejectable Force, we check for each halfspace that determines the polytope to check if the worst case force direction would violate the associated constraint at that value of $\rho$.
Given half-space representations $a_i F \leq b_i$, where $i$ indices the halfspaces of the RFP, we know the worst case force is in the direction of $a_i$ (see \cite{boyd2004convex}).
Hence we solve the following optimisation problem using enumeration:
\begin{equation}
  \max_{\rho} \boldsymbol{\rho} \quad\quad \text{s.t.} \quad \rho||a_i|| \leq b_i \quad \forall i =
  \max_i(b_i||a_i||^{-1})
\end{equation}
The Fibonnacci approximation still requires a conversion from vertex to halfspace representation. The computation time depends on the number of vertices sampled.
We use 1024 samples for a high quality approximation.

The affine and quadratic approximations of the SUF do not compute the RFP explicitly.
Instead, they find the worst case disturbance force (similar as above), while simultaneously solving for an optimal control law determining how the joint-torques and ground reaction forces change with the disturbance force.
As the true (nonlinear) optimal control law cannot be computed efficiently, the two approximations assume an affine and quadratic control law respectively~\cite{zhen2017computing}.
The optimisation problems deviate slightly from those in \cite{zhen2017computing} to simplify handling of the equality constraints for this specific scenario, and to search for the largest sphere centred around the origin, rather than around an arbitrary point.

First reparametrise the control equations:
\begin{equation*}
  \begin{bmatrix}
    F & \tau & \lambda
    \end{bmatrix}^{\top} =
    \begin{bmatrix}
      I & 0 \\ -W^{+} Je^{\top} & N  
    \end{bmatrix} \begin{bmatrix}F \\ \delta Q\end{bmatrix} + \begin{bmatrix} 0 \\ W^{+} d\end{bmatrix}
\end{equation*}
where $\delta Q$ are combinations of joint-torques and ground reaction forces in the null-space of the dynamics equation, which are solved using the matrix $W = \begin{bmatrix}B & J_f^{\top} & J_p^{\top}\end{bmatrix}$, the matrix $N$ is a basis for the nullspace of $W$ and the Moore-Penrose pseudo-inverse is indicated by a $+$.

The affine approximation optimises an affine control law from disturbance to reaction forces and torques:
\begin{equation}
  \delta Q = \delta Q_0 + VF
\end{equation}
where $\delta Q_0$ are nominal joint torques and ground reaction forces, and  $V$ is a gain matrix. These parameters are optimised along with the size of the SUF ($\rho$), via the conical quadratic program:
\begin{equation}
  \max_{\rho, \delta Q_0, V} \rho \quad \text{s.t.} \; \begin{bmatrix}0 & \delta Q_0\end{bmatrix}a_i + \left|\begin{bmatrix} \rho I & V^{\top}\end{bmatrix}a_i\right|\leq b_i \; \forall i  
\end{equation}
the constraint coefficients in $a_i$ and $b_i$ are taken from Eqs. \ref{eq:constraints}-\ref{eq:torque_bounds}. The quadratic approximation uses a quadratic control law. The resulting semi-definite program is included in the Appendix.

\begin{figure}[b]
    \centering
    \begin{overpic}[trim={0 0 0 0},clip,width=0.95\columnwidth]{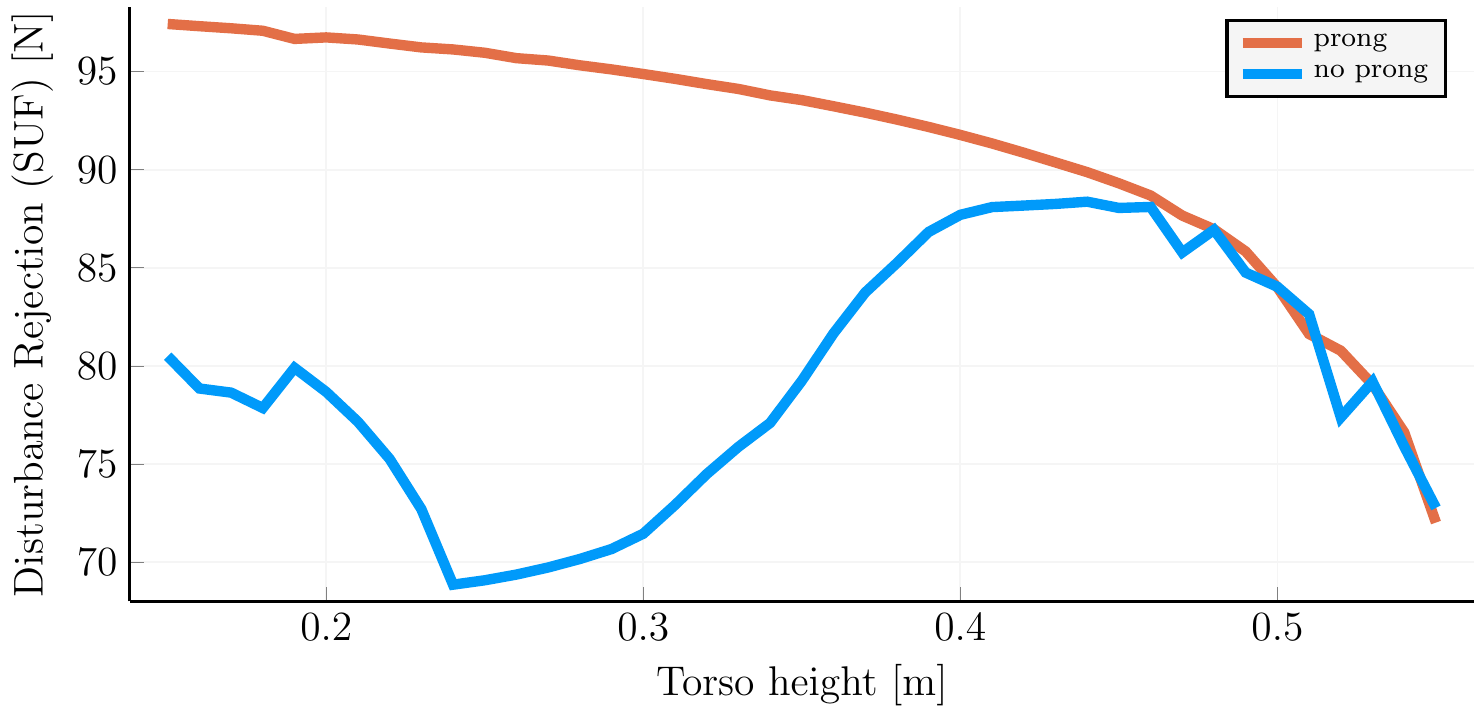}
        \put(160,50){\Huge \faLongArrowRight}
        \put(140,40){\small Approaching Max}
        \put(150,30){\small Base Height}
    \end{overpic}
    \caption{The SUF for each height of the base. The foot locations are optimized for each height.}
    \label{fig:optimized_per_height}
\end{figure}
\section{Simulation Experiments}
\label{sec:simulation_results}

This section first compares the proposed approximations on computational efficiency and accuracy. 
Then the prong optimisation problem from Section \ref{sec:prong_design} is solved using the affine approximation of the SUF.

\subsection{Comparing Approximation Methods}

This section compares the computation time and accuracy of all computation methods: exact, single direction \cite{orsolino2018application}, Fibonacci, affine and quadratic. The simulations were implemented with \texttt{Julia}'s libraries for rigid body dynamics \cite{koolen2019julia} and optimisation \cite{dunning2017jump, odonoghue2017splitting}.
The code ran on a PC with Intel Core i7-7830x processor and 32Gb of memory.

To compare the approximations the SUF was computed for random robot configurations from two scenarios:
\begin{enumerate*} 
    \item tele-operation scenario, similar to \cite{xin2019bounded}, in which there is no arm attached to the robot, three legs are on the ground, and the remaining leg is used as end-effector
    \item a scenario with an arm attached to the robot functioning as end-effector, and all four legs of the robot in contact with the ground (see Fig \ref{fig:optimal_no_prong}).
\end{enumerate*}

The results, shown in Table \ref{tab:suf_computation_time}, affirm the slow computation of the exact method.
The affine and quadratic approximations are faster than the Fibonacci approach.
The quadratic approximation scales less well to the arm-attached scenario, due to the number of parameters in the quadratic term of the control law.
The single direction approach is clearly fastest.

\begin{figure}[t]
    \centering
    \begin{subfigure}{1.0\columnwidth} 
        \includegraphics[width=\textwidth]{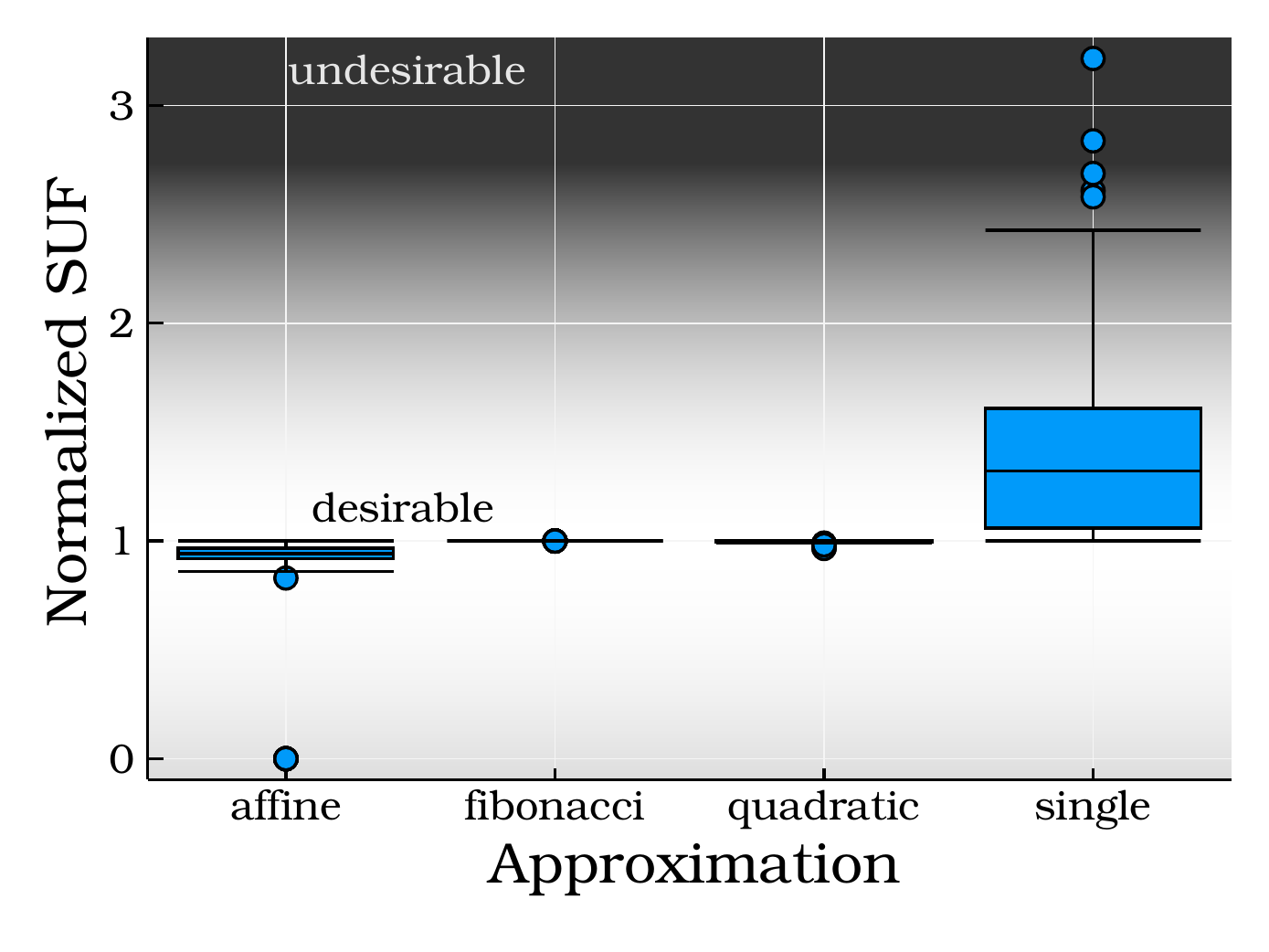}
    \end{subfigure}
    \caption{Boxplot of relative SUF size in each approximation. The background color indicates the desirability: accurate approximations and underestimations are more desirable.}
    \label{fig:compare_approx}
\end{figure}

Figure \ref{fig:compare_approx} conveys the quality of the approximations, by comparing the sizes of their SUFs relative to exact computation. Only the arm-attached scenario is shown to highlight the potential differences in quality, as those are larger in that scenario.
The quadratic and Fibonacci approximations are very close to exact, with the affine approach resulting in slight underapproximations.
The affine approximation failed to converge once due to numerical issues.
Note these three approximations are conservative.
The single-direction approximation is shown to have poor accuracy.
Furthermore, this approach overestimates of the SUF, which is undesirable for robustness analysis.
Due to the favourable trade-off between conservativity, accuracy and computation time, the affine approximation is used in the remainder of this paper.

\subsection{Optimizing Robust Body-Ground Contact}

We optimised the prong length and placement, in order to maximise the SUF for a fixed end-effector position, see Eq. \ref{eq:optimization}. The results of this optimisation are shown for scenarios with and without prong in Figure \ref{fig:optimal_no_prong}.
Shown are the optimal configuration of the robot, the resulting rejectable force polytope, and a sphere with the SUF as radius. The forces are scaled using a `stiffness' of \SI{1000}{\newton\per\metre}. The polytopes and spheres are translated, such that their origin (0 disturbance force) is at the end-effector.
\begin{table*}[t]
\centering
\vspace{0.3cm}
  \caption{Computation time of SUF-approximations in \si{\milli\second}}
  \label{tab:suf_computation_time}
  \begin{tabular}{cccccc}
  \toprule
  Task & Exact &  Fibonacci & Affine & Quadratic & Single\\
  \midrule
   teleoperation  & 4301 &  588.8 & 13.8 & 84.169 & 4.65 \\
   manipulation   & 37747 & 727.0 & 17.49 & 709.1 & 4.93 \\
  \bottomrule
  \end{tabular}
\end{table*}
The minimal non-rejectable forces are \SI{88}{\newton} and \SI{96}{\newton} respectively.

To show the efficacy of prongs, we also found the SUF given a body height, optimising only the foot locations, see Figure \ref{fig:optimized_per_height}.
Slight noise is caused by approximations in the IK algorithm. 
When the \prongs~ are attached, the SUF is pointed upwards, and is limited by the unilaterality conditions. The prong's length has little effect in this direction, so does not effect the SUF.
Therefore, the prong length can be decided by other considerations: ground clearance and a minimum height from the base. 
The prongs also have little effect on the SUF when the torso height is larger, as the robot legs are then close to their singular position, which limits joint torques.
However, as this singularity comes with mobility and control issues, such heights are undesirable.
For more practical torso heights we see that the prongs provide a benefit of up to 35\%.

\begin{figure}[t!]
    \centering
    \includegraphics[trim={0 0 0 18cm},clip,width=0.9\columnwidth]{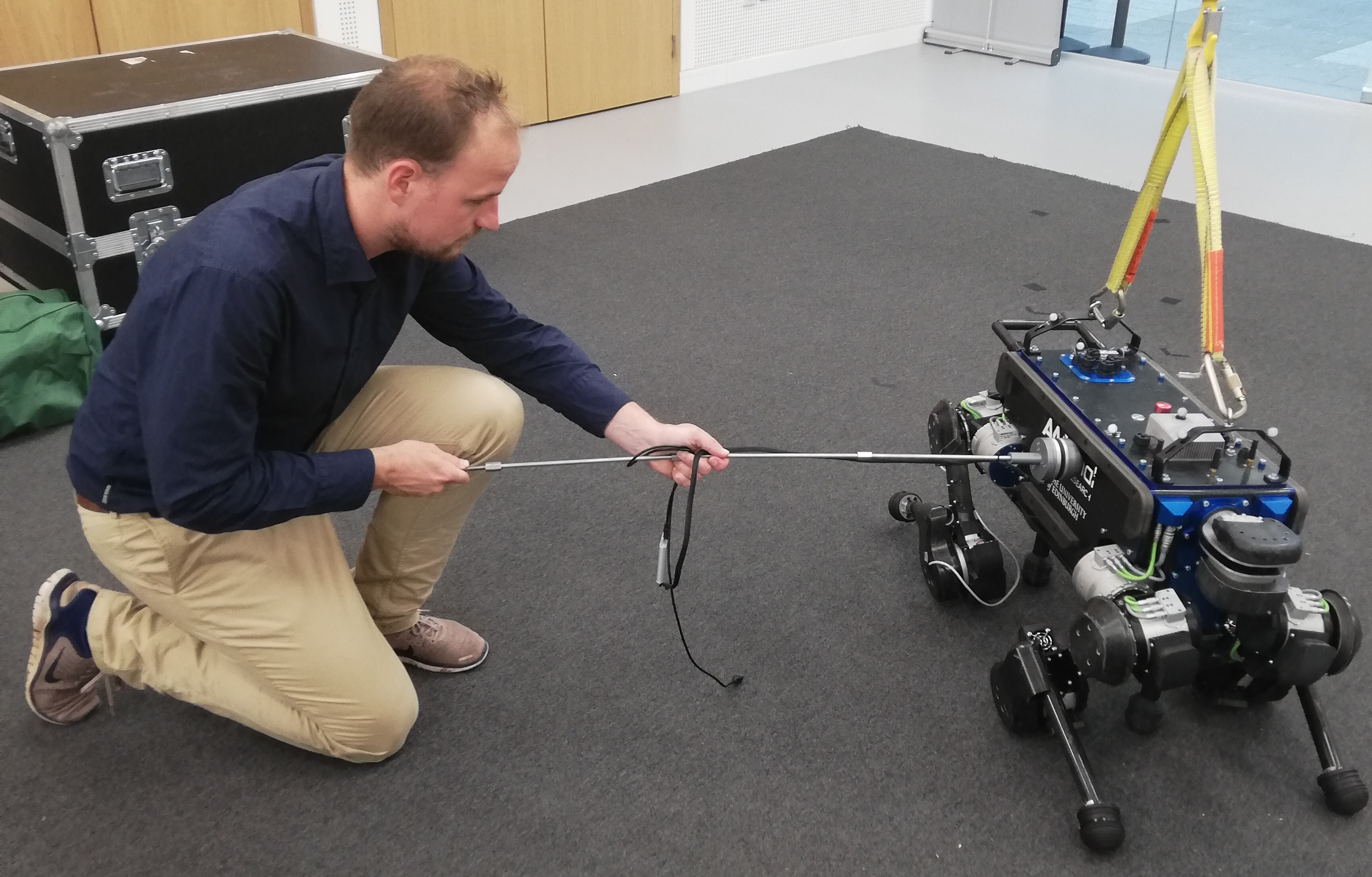}
    \caption{The push recovery experimental setup. The robot is pushed using a stick with a force/torque sensor.}
    \label{fig:pushing_robot}
\end{figure}

\section{Controller Design}
\label{sec:control}

The whole-body-control of ANYmal uses the well established hierarchical QP paradigm~\cite{de2010feature}.
To ensure the robustness of this control paradigm, some additions have been proposed recently. 
For example, the techniques from \cite{del2016robustness} aim to improve robustness against joint tracking errors.
In this paper, we follow the hierarchical QP framework by combining the foot contact constraints and the prong contact constraints into a single augmented contact Jacobian. 
As such, the prong and foot contacts are considered in the same way and their forces are optimised simultaneously. 

At each time-step, as part of the QP, the controller minimises an error between the desired and actual task-space accelerations space: $||\ddot{x} - \ddot{x}_\text{d}||_w^2$, with the desired accelerations based on the error $e$ in the task space position.
If for $\ddot{x} = \ddot{x}_{d}$ the kinematic constraints do not hold, i.e. the system is overconstrained with respect to its desired movement, minimising the acceleration error might lead to unstable behaviour.
To ensure a solution that stabilises the robot, we use a whole-body controller consisting of the following five hierarchical layers, each solving a QP. Each layer ensures that the optimality conditions of the previous layers are satisfied, i.e., it optimises in the nullspace of the previous layers.
\begin{enumerate}
    \item Dynamic feasibility: finds any feasible solution for the dynamic constraints, Eqs. \ref{eq:eom}-\ref{eq:torque_bounds}.
    \item Torso angular acceleration tracking: minimises the error between the desired angular acceleration and the executed angular acceleration.
    \item Torso translational acceleration tracking: when the prongs contact the ground, this layer has no effect on the outcome, as translational acceleration is not in the available nullspace. This prevents unstable behaviour.
    \item Swing foot acceleration tracking 
    \item Torque minimisation: minimises the sum of squared motor torques, in order to reduce energy consumption.
\end{enumerate}
The implementation also reduces computational load via the trick from \cite{herzog2016momentum} to avoid directly computing the torques.

\begin{figure}[t]
\vspace{0.3cm}
    \centering
    \begin{overpic}[scale=1,trim={0 0 0 0},clip,width=0.9\columnwidth]{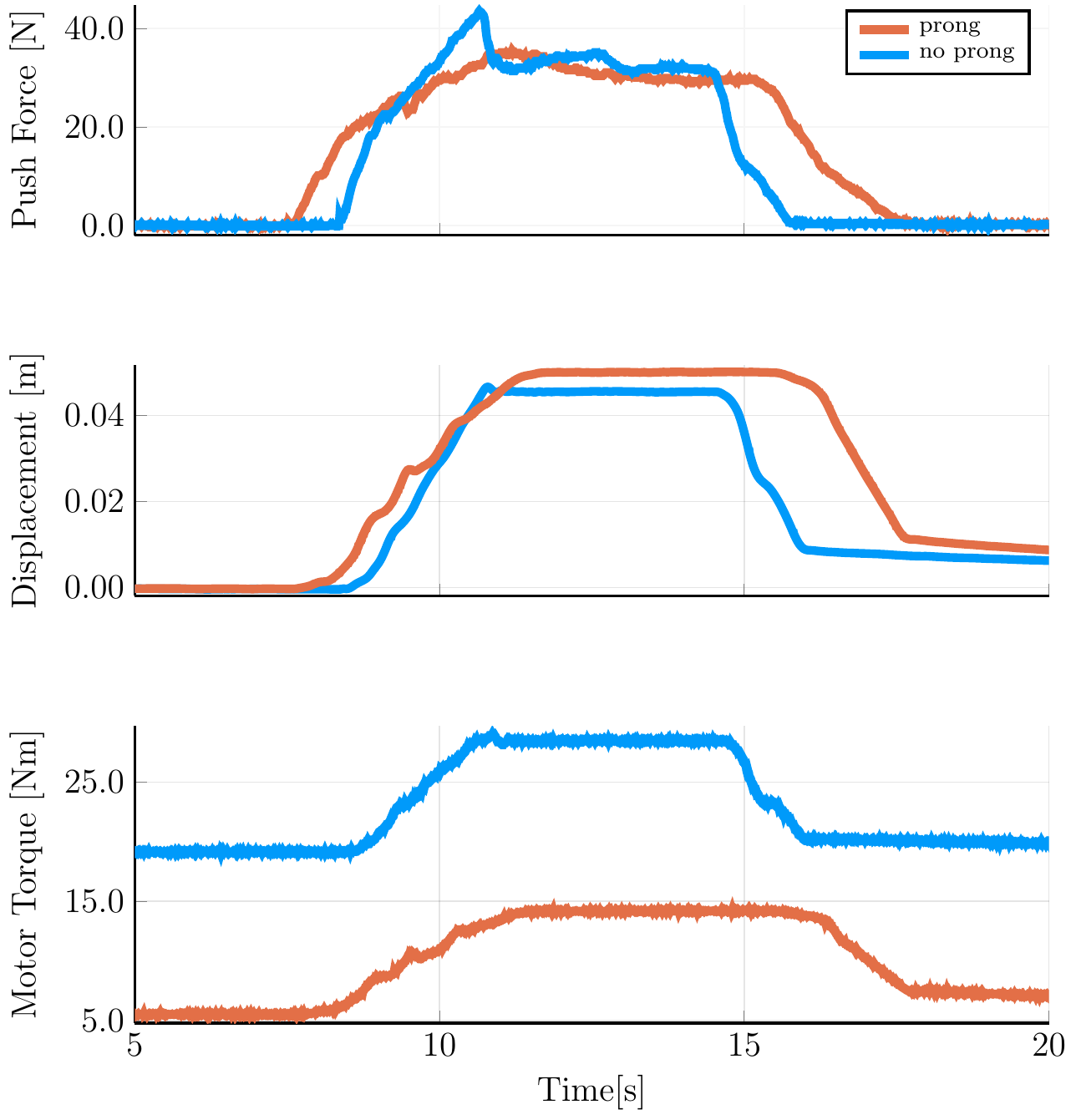}
        \put(35,30){\Huge \faLongArrowDown}
        \put(45,46){\small Torque Reduction}
    \end{overpic}
    \caption{Comparison of torso displacement, maximum push force and required motor torque (left front knee) when pushing the robot with (red) and without (blue) prongs.}
    \label{fig:pushing_results}
    \vspace{-5mm}
\end{figure}

\section{Hardware Experiments}
\label{sec:experiments}

To validate the use of prongs, we perform three experiments.
\begin{enumerate*}
    \item Pushing the robot to assess the joint-torques.
    \item Clearing out an obstacle with the robots' conventional legs, freed for this task by the prongs.
    \item Lifting a box with two side legs, to establish the versatility of the controller.
\end{enumerate*}
\newcommand{\pushtimelinewidth}{0.329}
\newcommand{\pushtrimtop}{5cm}
\newcommand{\pushtrimbottom}{7cm}
\newcommand{\pushtrimleft}{9cm}
\begin{figure*}[t]
\vspace{0.5cm}
    \centering
    \begin{subfigure}{\textwidth}
        \includegraphics[width=\textwidth]{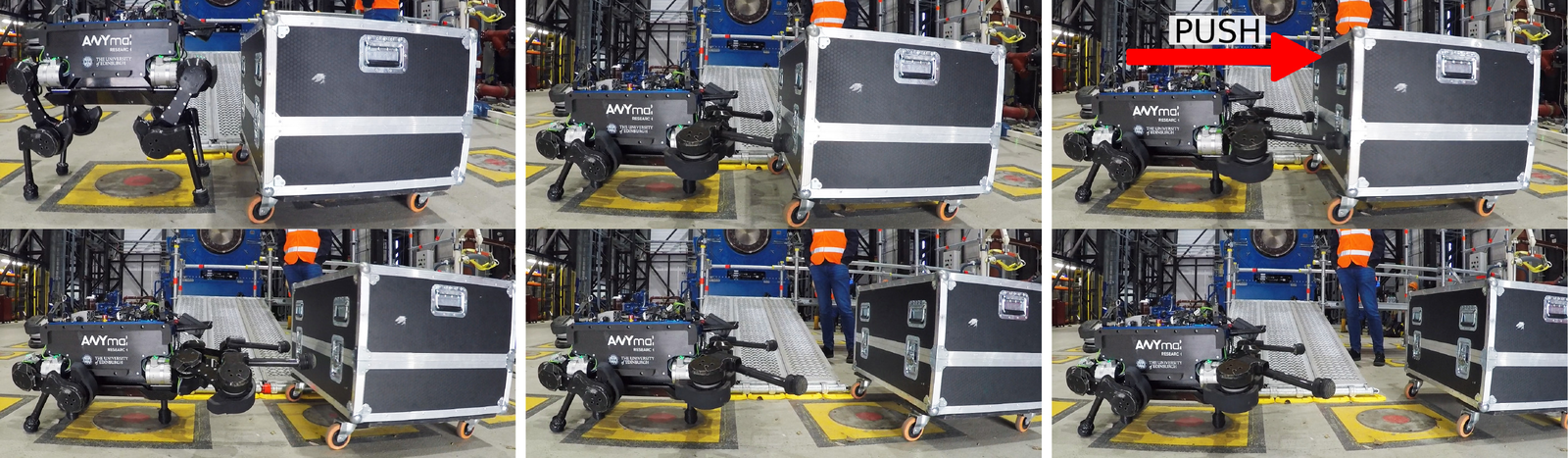}
    \end{subfigure}
    \centering
    \caption{The robot leaning on both prongs to push a box away. From top left: From a standing position (1), the robot lowers itself onto the prongs (2), which enables the front legs to be lifted from the ground. These front legs are used to push the box (3), which is pushed out of the way (4-6), clearing space for the robot to move into.}
    \label{fig:box_push1}
\end{figure*}

\renewcommand{\pushtimelinewidth}{0.19}
\begin{figure*}[t]
    \centering
    \begin{subfigure}{\textwidth}
        \includegraphics[width=\textwidth]{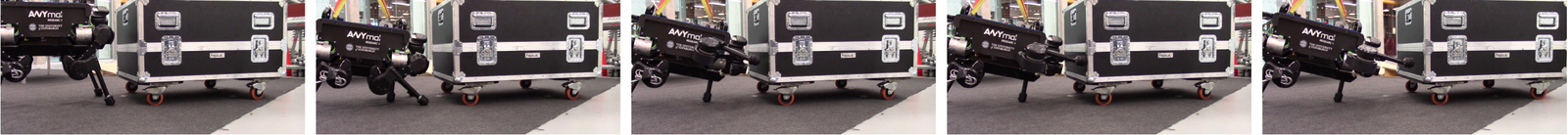}
    \end{subfigure}
    \centering
    \caption{The robot leaning on both prongs to push a box away. From top left: From a standing position (1), the robot lowers itself onto the prongs (2), which enables the front legs to be lifted from the ground. These front legs are used to push the box (3), which is pushed out of the way (4-6), clearing space for the robot to move into.}
    \label{fig:box_push2}
\end{figure*}

\newcommand{\trimright}{13cm}
\newcommand{\trimleft}{15cm}
\newcommand{\trimbottom}{6cm}
\newcommand{\lifttimelinewidth}{0.195}

\begin{figure*}
    \centering
    \begin{subfigure}{\textwidth}
        \includegraphics[width=\textwidth]{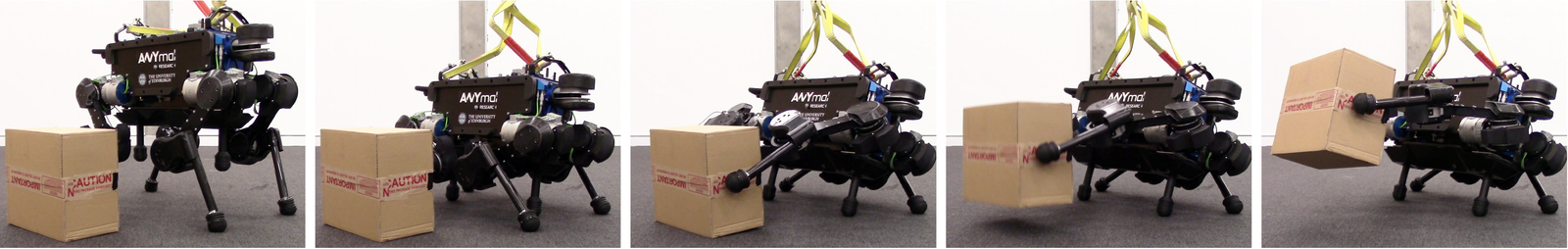}
    \end{subfigure}
    \caption{The support provided by the prongs is used to free two of the legs for a manipulation task (i.e., lifting a box)}
    \label{fig:box_lift1}
\end{figure*}

\renewcommand{\trimright}{0cm}
\renewcommand{\trimleft}{3cm}
\renewcommand{\trimbottom}{0cm}
\renewcommand{\lifttimelinewidth}{0.195}

\begin{figure*}
    \centering
    \begin{subfigure}{\textwidth}
        \includegraphics[width=\textwidth]{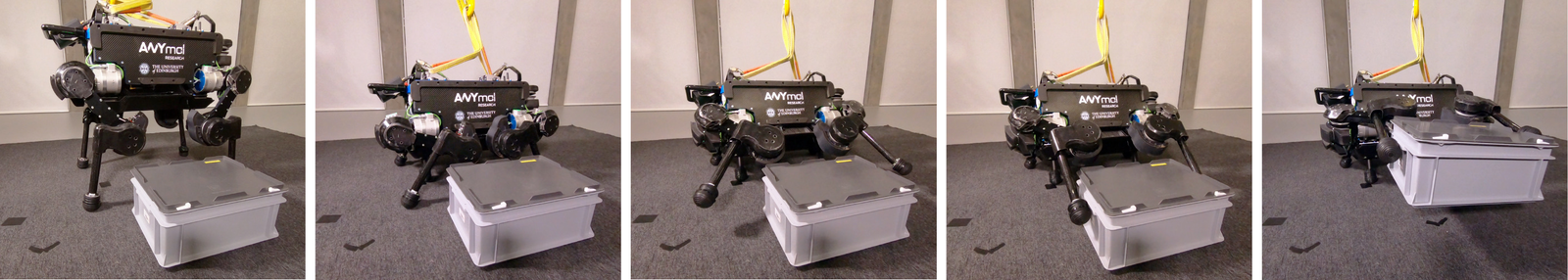}
    \end{subfigure}
    \caption{Motion stills of a second box lifting experiment, using a differently sized box.}
    \label{fig:box_lift2}
\end{figure*}

\subsection{Push Recovery}

To verify that our controller is able to use the prongs to enhance the robustness of the robot, we push the robot with a rod equipped with a force-torque sensor. The experiment is shown in Figure \ref{fig:pushing_robot}.
In the experiment, we push the robot horizontally on the middle of the base. The robot is in the same configuration for both experiments, which is the default standing configuration of the robot-platform with the torso height lowered to the height of the prongs. The force is gradually increased up to approximately 30 N. After holding this force for 5 seconds, the force is reduced to 0.

Figure \ref{fig:pushing_results} shows the base displacement, disturbance force and motor torque during the experiment.
The key result is found by comparing the effective compliance and the amplification factor between disturbance force and motor torque during the period of maximum push force.
The motor torque is significantly lower with prong, despite a slightly stiffer torso behaviour.
These reduced knee-joint torques result in a capacity to reject larger disturbance forces.

\subsection{Obstacle Clearance}

The second experiment shows robot's ability to perform basic manipulation using the \prongs.
Standard quadrupeds would be unable to perform manipulation with more than one leg, as they are required for standing. 
The \prongs~take over responsibility for standing, freeing the legs for manipulation.
Using the free-gait motion description library~\cite{fankhauser2016free}, we generate a sequence of body and end-effector targets, such that the robots pushes an obstacle away.

The resulting motion can be seen in Figures \ref{fig:box_push1} and \ref{fig:box_push2}.
Note that, by necessity, the legs of a quadruped are equipped with relatively strong motors, which makes them well suited for obstacle clearing tasks such as this.
Performing such a task is only possible when relying on body-ground contact.
Enhancing the capabilities of the robot to allow obstacle clearing makes them more versatile in rough terrain.

\subsection{Lifting Box}

Experiment three shows controlled torso mobility while the \prongs~contact the ground, and further manipulation with the legs.
The robot lowers on its prongs, and leans to its right side, freeing the left side legs for manipulation.
The legs are controlled in task-space with low-gains, allowing basic dual arm manipulation: picking up a box. 
Figures \ref{fig:box_lift1} and \ref{fig:box_lift2} shows snapshots of this motion, lifting two different boxes. 
This and the previous experiment are captured in the accompanying video.

When lifting the left-side legs, the desired ground-contact force changes quickly, which can result in jerky motions.
To counteract such motions, we enforce a smooth contact force.
Future work on control should focus on the capability of the robot to track more rapid transitions.

\section{Discussion}
\label{sec:discussion}

The improvement in robustness and manipulation capabilities introduced by the body-ground contact opens several applications, improving the intervention capabilities in robots deployed for exploration and monitoring.
However, body-ground contact needs further research before quadrupeds are ready for such applications.

There are three main areas of improvement for future work.
First is extending the affine and quadratic approximations of the SUF.
Knowledge about the distribution of contact forces can be used to bias the robustness measure towards more likely disturbances.
With further improvements to the computational efficiency, these approximations could be incorporated in real-time planning and control.
Second is incorporating body-ground contact in motion planning.
Making decisions about when to use the prongs (for robustness) and when not (for speed), maximising the utility of the prongs is a challenging problem due to the intermittent nature of the contact.
Third is applying prongs to body-environment contact with other robot morphologies.
Two examples are: a quadruped increasing its robustness by leaning sideways on a wall with its body, and a robot arm increasing its accuracy by contacting a table with a prong on its elbow.

\section{Conclusion}
\label{sec:conclusion}

This paper studied the use of \prongs~ to enable body-ground contact in quadrupedal robots.
We showed that using \prongs~ increases the robustness of the robot, as measured by its ability to reject forces applied to the end-effector, by up to 35 \%, largely independent on prong length.

We applied an optimisation-based whole-body controller that handles the constrained body motion resulting from body-ground contact.
On the hardware, we verified the increased robustness in the form of push resistance with limited motor torques.
We also showed obstacle clearance and basic object manipulation, two capacities added by the prongs freeing the legs from their body-support task.

\FloatBarrier

\appendix
\section{Quadratic Inverse Dynamics Rule}

To compute the SUF, it is also possible to use a quadratic inverse dynamics law.
Optimising this law for maximal force rejection, is a semi-definite program.
The program is detailed below, adapted from \cite{zhen2017computing}.
\begin{align}
    &\max_{\rho, \delta Q_0, V, W, \zeta, \xi} \rho \quad \text{s.t.} \quad\quad \zeta \leq \boldsymbol{b}\\
    & \hspace{10em}  \xi \geq 0 \\
    & \small{\begin{pmatrix} \zeta_i - \xi_i - \boldsymbol{a}_i^{\top}\begin{pmatrix} 0 \\ \delta Q_0 \end{pmatrix} &
        -\frac{1}{2}\boldsymbol{a}_i^{\top}\begin{pmatrix}\rho I \\ V\end{pmatrix}  \\
    -\frac{1}{2}\begin{pmatrix}\rho I \\ V\end{pmatrix}^{\top}\boldsymbol{a}_i
 & \xi_i - \sum_j^{n_2}a_{ij+n_1}W_j\end{pmatrix}} \succeq 0 \: \forall \: i
\end{align}
Here the quadratic term of the inverse dynamics law is $W$,
and we introduced two helper variables $\xi$ and $\zeta$.

\balance
\bibliography{references.bib}

\begin{thebibliography}{10}
\providecommand{\url}[1]{#1}
\csname url@samestyle\endcsname
\providecommand{\newblock}{\relax}
\providecommand{\bibinfo}[2]{#2}
\providecommand{\BIBentrySTDinterwordspacing}{\spaceskip=0pt\relax}
\providecommand{\BIBentryALTinterwordstretchfactor}{4}
\providecommand{\BIBentryALTinterwordspacing}{\spaceskip=\fontdimen2\font plus
\BIBentryALTinterwordstretchfactor\fontdimen3\font minus
  \fontdimen4\font\relax}
\providecommand{\BIBforeignlanguage}[2]{{%
\expandafter\ifx\csname l@#1\endcsname\relax
\typeout{** WARNING: IEEEtran.bst: No hyphenation pattern has been}%
\typeout{** loaded for the language `#1'. Using the pattern for}%
\typeout{** the default language instead.}%
\else
\language=\csname l@#1\endcsname
\fi
#2}}
\providecommand{\BIBdecl}{\relax}
\BIBdecl

\bibitem{farnioli2015optimal}
E.~Farnioli, M.~Gabiccini, and A.~Bicchi, ``Optimal contact force distribution
  for compliant humanoid robots in whole-body loco-manipulation tasks,'' in
  \emph{Int. Conf. on Robotics and Automation}, 2015.

\bibitem{abi2019torque}
F.~Abi-Farraj, B.~Henze, C.~Ott, P.~R. Giordano, and M.~A. Roa, ``Torque-based
  balancing for a humanoid robot performing high-force interaction tasks,''
  \emph{Robotics and Automation Letters}, 2019.

\bibitem{kudruss2015optimal}
M.~Kudruss, M.~Naveau, O.~Stasse, N.~Mansard, C.~Kirches, P.~Soueres, and
  K.~Mombaur, ``Optimal control for whole-body motion generation using
  center-of-mass dynamics for predefined multi-contact configurations,'' in
  \emph{Int. Conf. on Humanoid Robots}, 2015.

\bibitem{henze2016passivity}
B.~Henze, M.~A. Roa, and C.~Ott, ``Passivity-based whole-body balancing for
  torque-controlled humanoid robots in multi-contact scenarios,'' \emph{Int. J.
  of Robotics Research}, 2016.

\bibitem{carpentier2018multicontact}
J.~Carpentier and N.~Mansard, ``Multicontact locomotion of legged robots,''
  \emph{T. on Robotics}, 2018.

\bibitem{koolen2016design}
T.~Koolen, S.~Bertrand, G.~Thomas, T.~De~Boer, T.~Wu, J.~Smith, J.~Englsberger,
  and J.~Pratt, ``Design of a momentum-based control framework and application
  to the humanoid robot atlas,'' \emph{Int. J. of Humanoid Robotics}, 2016.

\bibitem{smith2012dual}
C.~Smith, Y.~Karayiannidis, L.~Nalpantidis, X.~Gratal, P.~Qi, D.~V.
  Dimarogonas, and D.~Kragic, ``Dual arm manipulation—a survey,''
  \emph{Robotics and Autonomous systems}, 2012.

\bibitem{xin2018model}
G.~Xin, H.-C. Lin, J.~Smith, O.~Cebe, and M.~Mistry, ``A model-based
  hierarchical controller for legged systems subject to external
  disturbances,'' in \emph{Int. Conf. on Robotics and Automation}, 2018.

\bibitem{henze2017multi}
B.~Henze, A.~Dietrich, M.~A. Roa, and C.~Ott, ``Multi-contact balancing of
  humanoid robots in confined spaces: Utilizing knee contacts,'' in \emph{Int.
  Conf. on Intelligent Robots and Systems}, 2017.

\bibitem{trkov2019bipedal}
M.~Trkov, K.~Chen, and J.~Yi, ``Bipedal model and hybrid zero dynamics of human
  walking with foot slip,'' \emph{J. of Computational and Nonlinear Dynamics},
  2019.

\bibitem{specian2018robotic}
A.~Specian, C.~Mucchiani, M.~Yim, and J.~Seo, ``Robotic edge-rolling
  manipulation: A grasp planning approach,'' \emph{Robotics and Automation
  Letters}, 2018.

\bibitem{van2015learning}
H.~Van~Hoof, T.~Hermans, G.~Neumann, and J.~Peters, ``Learning robot in-hand
  manipulation with tactile features,'' in \emph{Int. Conf. on Humanoid
  Robots}, 2015.

\bibitem{fazeli2019see}
N.~Fazeli, M.~Oller, J.~Wu, Z.~Wu, J.~Tenenbaum, and A.~Rodriguez, ``See, feel,
  act: Hierarchical learning for complex manipulation skills with multisensory
  fusion,'' \emph{Science Robotics}, 2019.

\bibitem{hutter2014quadrupedal}
M.~Hutter, H.~Sommer, C.~Gehring, M.~Hoepflinger, M.~Bloesch, and R.~Siegwart,
  ``Quadrupedal locomotion using hierarchical operational space control,''
  \emph{Int. J. of Robotics Research}, 2014.

\bibitem{shirafuji2016mechanism}
S.~Shirafuji, Y.~Terada, and J.~Ota, ``Mechanism allowing a mobile robot to
  apply a large force to the environment,'' in \emph{Int. Conf. on Intelligent
  Autonomous Systems}, 2016.

\bibitem{kiyota20063d}
T.~Kiyota, N.~Sugimoto, and M.~Someya, ``3d-free rescue robot system,'' in
  \emph{Int. Conf. on Robotics and Automation}, 2006.

\bibitem{jeng2010outrigger}
S.-L. Jeng, C.-F. Yang, and W.-H. Chieng, ``Outrigger force measure for mobile
  crane safety based on linear programming optimization\#,'' \emph{Mechanics
  Based Design of Structures and Machines}, 2010.

\bibitem{oh2017technical}
P.~Oh, K.~Sohn, G.~Jang, Y.~Jun, and B.-K. Cho, ``Technical overview of team
  drc-hubo@ unlv's approach to the 2015 darpa robotics challenge finals,''
  \emph{J. of Field Robotics}, 2017.

\bibitem{bjelonic2018skating}
M.~Bjelonic, C.~D. Bellicoso, M.~E. Tiryaki, and M.~Hutter, ``Skating with a
  force controlled quadrupedal robot,'' in \emph{Int. Conf. on Intelligent
  Robots and Systems}, 2018.

\bibitem{saab2018robotic}
W.~Saab, W.~S. Rone, and P.~Ben-Tzvi, ``Robotic tails: a state-of-the-art
  review,'' \emph{Robotica}, 2018.

\bibitem{thibodeau2006static}
B.~J. Thibodeau, P.~Deegan, and R.~Grupen, ``Static analysis of contact forces
  with a mobile manipulator,'' in \emph{Int. Conf. on Robotics and Automation},
  2006.

\bibitem{del2016robustness}
A.~Del~Prete and N.~Mansard, ``Robustness to joint-torque-tracking errors in
  task-space inverse dynamics,'' \emph{T. on Robotics}, 2016.

\bibitem{escande2014hierarchical}
A.~Escande, N.~Mansard, and P.-B. Wieber, ``Hierarchical quadratic programming:
  Fast online humanoid-robot motion generation,'' \emph{Int. J. of Robotics
  Research}, 2014.

\bibitem{bicchi1993force}
A.~Bicchi, ``Force distribution in multiple whole-limb manipulation,'' in
  \emph{Int. Conf. on Robotics and Automation}, 1993.

\bibitem{orsolino2018application}
R.~Orsolino, M.~Focchi, C.~Mastalli, H.~Dai, D.~G. Caldwell, and C.~Semini,
  ``Application of wrench-based feasibility analysis to the online trajectory
  optimization of legged robots,'' \emph{Robotics and Automation Letters},
  2018.

\bibitem{ferrolho2019comparing}
H.~Ferrolho, W.~Merkt, C.~Tiseo, and S.~Vijayakumar, ``Comparing metrics for
  robustness against external perturbations in dynamic trajectory
  optimization,'' \emph{arxiv}, 2019.

\bibitem{xin2019bounded}
G.~Xin, J.~Smith, D.~Rytz, W.~Wolfslag, H.-C. Lin, and M.~Mistry, ``Bounded
  haptic teleoperation of a quadruped robot's foot posture for sensing and
  manipulation,'' in \emph{Int. Conf. on Robotics and Automation}, 2020.

\bibitem{boyd2004convex}
S.~Boyd and L.~Vandenberghe, \emph{Convex optimization}.\hskip 1em plus 0.5em
  minus 0.4em\relax Cambridge University Press, 2004.

\bibitem{zhen2017computing}
J.~Zhen and D.~Den~Hertog, ``Computing the maximum volume inscribed ellipsoid
  of a polytopic projection,'' \emph{INFORMS J. on Computing}, 2017.

\bibitem{koolen2019julia}
T.~Koolen and R.~Deits, ``Julia for robotics: Simulation and real-time control
  in a high-level programming language,'' in \emph{Int. Conf. on Robotics and
  Automation}, 2019.

\bibitem{dunning2017jump}
I.~Dunning, J.~Huchette, and M.~Lubin, ``Jump: A modeling language for
  mathematical optimization,'' \emph{SIAM Review}, 2017.

\bibitem{odonoghue2017splitting}
B.~O'Donoghue, E.~Chu, N.~Parikh, and S.~Boyd, ``Conic optimization via
  operator splitting and homogeneous self-dual embedding,'' \emph{J. of
  Optimization Theory and Applications}, June 2016.

\bibitem{de2010feature}
M.~De~Lasa, I.~Mordatch, and A.~Hertzmann, ``Feature-based locomotion
  controllers,'' in \emph{ACM Transactions on Graphics (TOG)}, 2010.

\bibitem{herzog2016momentum}
A.~Herzog, N.~Rotella, S.~Mason, F.~Grimminger, S.~Schaal, and L.~Righetti,
  ``Momentum control with hierarchical inverse dynamics on a torque-controlled
  humanoid,'' \emph{Autonomous Robots}, 2016.

\bibitem{fankhauser2016free}
P.~Fankhauser, C.~D. Bellicoso, C.~Gehring, R.~Dub{\'e}, A.~Gawel, and
  M.~Hutter, ``Free gait—an architecture for the versatile control of legged
  robots,'' in \emph{Int. Conf. on Humanoid Robots}, 2016.

\end{thebibliography}
\bibliographystyle{IEEEtran}

\end{document}